# Clustering Filipino Disaster-Related Tweets Using Incremental and Density-Based Spatiotemporal Algorithm with Support Vector Machines for Needs Assessment

**Conference Paper** · April 2018

| CITATIONS | READS |
|---|---|
| 0 | 115 |

**7 authors**, including:

Charmaine Ponay
University of Santo Tomas
**2** PUBLICATIONS   **0** CITATIONS

Some of the authors of this publication are also working on these related projects:

Hiraya: Genre Classification of Filipino Fictions using K-Means Algorithm and Artificial Neural Networks   View project

# Clustering Filipino Disaster-Related Tweets Using Incremental and Density-Based Spatiotemporal Algorithm with Support Vector Machines for Needs Assessment

Ocean M. Barba, Franz Arvin T. Calbay, Angelica Jane S. Francisco, Angel Luis D. Santos, Charmaine S. Ponay

*Institute of Information and Computing Sciences*
*University of Santo Tomas, Manila*

{ocean.barba, franzarvin.calbay, angelicajane.francsico, angelluis.santos}.iics@ust.edu.ph, csponay@ust.edu.ph

## I. ABSTRACT

Social media has played a huge part on how people get informed and communicate with one another. It has helped people express their needs due to distress especially during disasters. Because posts made through it are publicly accessible by default, Twitter is among the most helpful social media sites in times of disaster. With this, the study aims to assess the needs expressed during calamities by Filipinos on Twitter. Data were gathered and classified as either disaster-related or unrelated with the use of Naïve Bayes classifier. After this, the disaster-related tweets were clustered per disaster type using Incremental Clustering Algorithm, and then sub-clustered based on the location and time of the tweet using Density-based Spatiotemporal Clustering Algorithm. Lastly, using Support Vector Machines, the tweets were classified according to the expressed need, such as "shelter", "rescue", "relief", "cash", "prayer", and "others".

After conducting the study, results showed that the Incremental Clustering Algorithm and Density-Based Spatiotemporal Clustering Algorithm were able to cluster the tweets with f-measure scores of 47.20% and 82.28% respectively. Also, the Naïve Bayes and Support Vector Machines were able to classify with an average f-measure score of 97% and an average accuracy of 77.57% respectively.

## II. INTRODUCTION

The Philippines has been plagued by different kinds of natural disasters such as typhoons, earthquakes and volcanic eruptions. This is due to its location along the Ring of Fire, a large Pacific Ocean Region where many of earth's volcanic eruptions and earthquakes occur [1].

In response to these disasters, the Philippine government formed an organization called the National Disaster Risk Reduction and Management Council (NDRRMC) in 2010 that is responsible for ensuring the protection and welfare of the people during a disaster or an emergency [2].

Assessment of the after effects is the first step in any emergency response caused by a disaster. This is where the utilization of social media platforms such as Twitter are used in which real-time propagation of information to a large group of users is supported.

With these in mind, the researchers created a Twitter-based system that can assess the needs of disaster hit areas from the collected tweets. The tweets were gathered from various government and news organizations, then the hashtags were saved and the tweets with similar hashtags were also collected. Such important information, if identified properly, can effectively help in improving the disaster awareness and disaster response of the general public.

## III. RELATED WORKS

One of the basis of the study was the work of [3] which deals with the classification of tweets according to its disaster-type and generation of template-based Filipino news. The preprocessing module used by [3] that includes tokenization and stop words removal Twas integrated to the system.

[4] proposed a similarity measure for the Incremental Clustering algorithm which is the Cluster Cohesion and Keyword Similarity. This was the basis for the similarity measure used in the Incremental Clustering module of the study.

[5] identified bursty areas of emergency situations using Incremental Density-Based Spatiotemporal Clustering Algorithm which was also used in the system to determine the location of different disasters.

## IV. ARCHITECTURE

Fig. 1 shows the system architecture of the system.

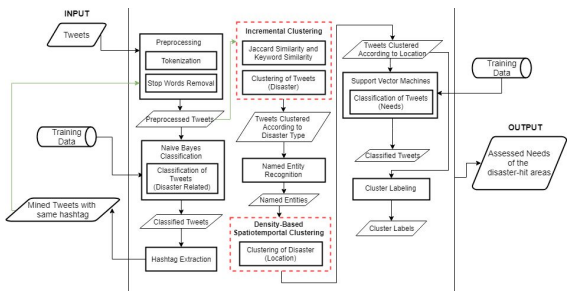

*Figure 1 System Architecture*

1. Preprocessing Module

This module cleans the tweet bodies by removing stop words. The tweets are first preprocessed in order to increase the accuracy of the classifiers and clustering algorithms.

1.1. Tokenization

The tokenization process converts tweet bodies into a list of words or tokens.

1.2. Stop Words Removal

This process omits words that are commonly used and are usually considered irrelevant to the study. For this sub-process, the researchers used the list of stop words which were lifted from [3].

2. Naïve Bayes Classification

The Naïve Bayes classifier was used for classifying the tweets as disaster-related or unrelated. The classifier is based on the Bayesian Theorem to calculate the probability of data $X$ belongs to class $C_i$ with the formula:

$$P(C_i \mid X) = \frac{P(X \mid Ci)\,P(Ci)}{P(X)}$$

The training data set consists of 3500 tweets lifted from the study of [3] in which 1542 tweets are disaster-related. The output for this module is a list of disaster-related tweets.

3. Hashtag Extraction

This module takes the hashtags from the tweet body of the disaster-related tweet. The tweets with similar hashtags were then collected real-time. A list of hashtags from disaster-related tweets is the output of the Hashtag Extraction module.

4. Incremental Clustering

This clustering algorithm was used for clustering tweets according to its disaster type. Jaccard Similarity and Keyword Similarity were the similarity measures used. The following are the formulas used in computing the similarity of the tweet with the cluster:

$$JaccardSim = \frac{D \cap C_j}{D \cup C_j}$$

$$KeywordSim = \sum_{i=1}^{n}(TF_{i,j} \times log(\frac{N}{CF_i}))$$

$$Sim = JaccardSim + KeywordSim$$

where $D$ represents the tokens in the tweet, $C_j$ represents the tokens in the cluster $j$, $TF_{i,j}$ is the number of occurence of term $i$ in in cluster $j$, $N$ is the number of clusters, and $CF_i$ is the number of Clusters containing the term $i$.

The computed similarity is compared to the threshold.value of 0.01 to check whether the tweet belongs to the cluster or not. If the tweet does not belong to any cluster, a new one will be created.

5. Named Entity Recognition

This module is responsible for tagging the locations contained on the tweets. It uses a list of locations to classify entities in the tweets. The algorithm returns the tweet with the tagged locations.

6. Density-Based Clustering

This module was used to cluster tweets according to location and time. The algorithm uses the location tag and the time of creation of the tweet to form clusters. The similarity measure used was Jaccard Similarity together with the Interarrival time. The Interarrival time is computed by the difference between the time the tweet was created and the time in the cluster.

A tweet is only added to a cluster if the similarity measure is greater than or equal to the threshold and the IAT is less than or equal to 7 days. The output of this module is a list of clusters formed according to the tagged location and time.

7. Support Vector Machines

The Support Vector Machines classifier was used to classify tweets according to the needs. The tokens are used as the features. The tweets are then transformed into vectors using TF-IDF (Term Frequency × Inverse Document Frequency) with the formula:

$$TF - IDF = TF_{i,j} \times log(\frac{N}{D_i})$$

where $TF_{i,j}$ is the frequency of term $i$ in tweet $j$, $N$ is the number of tweets and $D_i$ is the number of tweets that contain the term $i$. The algorithm then finds the hyperplane that separates different classes with the biggest margin.

The different classes for the needs are Rescue, Relief, Shelter, Cash, Prayer, or Others.

8. Cluster Labelling

The five most commonly used words per cluster were used in order to easily determine the topic of the cluster. The output of this module is a list of words that have the highest frequency per cluster.

V. EXPERIMENT

A. Evaluation of Naïve Bayes Classifier

The Naïve Bayes classifier was evaluated using Precision, Recall, and F-Measure. The results were shown in this part. Three test cases were prepared by the researchers. The first test case consists of 100% disaster-related tweets, the second test case contains 70% disaster-related tweets and the third test case contains 50% disaster-related tweets.

As seen in Table 3, it can be said that as the number of unrelated tweets increases, the computed precision also decreases since the number of false positives tweets increases. It also shows that the classifier classifies all disaster-related tweets correctly since the recall scores for the three test cases are 100%.

Table 3 Results of Naïve Bayes classifier

|  | Precision (%) | Recall (%) | F-Measure (%) |
|---|---|---|---|
| Test Case 1 | 100.00 | 100.00 | 100.00 |
| Test Case 2 | 95.72 | 100.00 | 97.81 |
| Test Case 3 | 87.26 | 100.00 | 93.19 |

B. Evaluation of Support Vector Machines

The Support Vector Machines was evaluated using Precision, Recall, and F-Measure. The results were shown in this section. The gamma used in the system is 0.01. Three test cases were prepared by the researchers. Each test case contains 210 tweets.

a. Gamma Values for SVM

As seen on Table 4, the gamma used in the system which is 0.01 is the gamma that gave the best accuracy.

Table 4 Accuracy for different gamma values

| GAMMA | Test 1 Accuracy (%) | Test 2 Accuracy (%) | Test 3 Accuracy (%) | Overall Accuracy (%) |
|---|---|---|---|---|
| 0.1 | 78.10 | 78.10 | 75.11 | 77.10 |
| 0.01 | 79.05 | 79.05 | 76.08 | 78.06 |
| 0.001 | 78.04 | 78.57 | 75.60 | 77.40 |
| 0.0001 | 78.51 | 79.03 | 75.58 | 77.71 |

b. Test Case 1

As shown in Table 5, the highest precision, recall, and F-measure among the needs is Others which is 97.67%, 89.36%, and 93.33% respectively. The need Shelter having the lowest precision with 55.56% and Cash having the lowest recall with 30%.

Table 5 Results of SVM for Test Case 1

| Needs | Precision (%) | Recall (%) | F-Measure (%) |
|---|---|---|---|
| Rescue | 67.57 | 75.76 | 71.73 |
| Relief | 69.77 | 83.33 | 75.96 |
| Shelter | 55.56 | 55.56 | 55.56 |
| Cash | 60.00 | 30.00 | 40.00 |
| Prayers | 66.67 | 73.68 | 70.00 |
| Others | 97.67 | 89.36 | 93.33 |

c. Test Case 2

As shown in Table 6, the need with the highest precision, recall, and F-measure is Others with 90.48%, 87.36%, and 88.89% scores respectively. The need rescue has the lowest precision with a 61.90% scores while Shelter has the lowest recall with a 62.50% score.

Table 6 Results of SVM for Test Case 2

| Needs | Precision (%) | Recall (%) | F-Measure (%) |
|---|---|---|---|
| Rescue | 61.90 | 76.47 | 79.05 |
| Relief | 79.55 | 74.47 | 76.92 |
| Shelter | 71.43 | 62.50 | 66.67 |
| Cash | 69.23 | 75.00 | 72.00 |
| Prayers | 76.92 | 71.43 | 74.07 |
| Others | 90.48 | 87.36 | 88.89 |

d. Test Case 3

As shown in Table 7, the need with the highest recall, and F-measure is Others with 91.55%, and 94.20% scores respectively. The need Prayers has the highest precision with a 100% score while Shelter had the lowest with a 57.14% score. The need Cash has the lowest recall score with 42.11%.

Table 7 Results of SVM for Test Case 3

| Needs | Precision (%) | Recall (%) | F-Measure (%) |
|---|---|---|---|
| Rescue | 68.00 | 79.07 | 73.12 |
| Relief | 63.27 | 70.45 | 66.67 |
| Shelter | 57.14 | 63.16 | 60.00 |
| Cash | 61.54 | 42.11 | 50.00 |
| Prayers | 100.00 | 69.23 | 81.89 |
| Others | 97.01 | 91.55 | 94.20 |

C. Evaluation of Incremental Clustering

The Incremental Clustering Algorithm was evaluated using Precision, Recall, and F-Measure. The results were shown in this part. The Precision, Recall, and F-Measure scores were computed by assigning each cluster to a class that is most frequent in the cluster and averaging the same measures of each cluster generated by the Incremental Clustering algorithm.

The researchers chose 0.01 among the other thresholds because it has the highest precision, recall, and F-measure scores with 69.31%, 35.79%, and 47.20% respectively as shown in Table 8.

Table 8 Results of Incremental Clustering for different Threshold values

| Threshold | Precision (%) | Recall (%) | F-measure (%) |
|---|---|---|---|
| 0.01 | 69.31 | 35.79 | 47.20 |
| 0.05 | 65.24 | 27.65 | 38.84 |
| 0.1 | 65.46 | 15.13 | 24.58 |
| 0.2 | 67.04 | 12.47 | 21.05 |
| 0.3 | 67.67 | 8.01 | 14.34 |
| 0.4 | 64.61 | 68.99 | 12.47 |

D. Evaluation of Density-Based Spatiotemporal Clustering Algorithm

The Density-Based Spatiotemporal Clustering Algorithm was evaluated using Precision, Recall, and F-Measure. The results were shown in this part. The Precision, Recall, and F-Measure scores were computed by assigning each cluster to a class that is most frequent in the cluster and averaging the same measures of each cluster generated by the Density-Based Spatiotemporal Clustering Algorithm.

The researchers chose 0.8 since it has the highest precision, recall, and F-measure among other thresholds with scores 84.54%, 80.13%, and 82.28% respectively as shown in Table 9.

Table 9 Results of Density-Based Spatiotemporal Clustering for different Threshold values

| Threshold | Precision (%) | Recall (%) | F-measure (%) |
|---|---|---|---|
| 0.1 | 82.06 | 77.14 | 79.52 |
| 0.2 | 82.29 | 78.50 | 80.44 |
| 0.4 | 84.16 | 79.73 | 81.89 |
| 0.6 | 84.44 | 79.89 | 82.10 |
| 0.8 | 84.54 | 80.13 | 82.28 |

VI. CONCLUSION AND RECOMMENDATIONS

After conducting the study, the researchers were able to conclude that the system was able to classify the tweets as disaster related and unrelated with the use of Naive Bayes Classifier, to cluster tweets with similar disaster types using Incremental Clustering Algorithm, to cluster tweets according to the location of the disaster using Density-Based

Spatiotemporal Clustering Algorithm, and to classify and assess the needs in each tweet using Support Vector Machines Classifier.

However, the researchers encountered several problems. Since the dataset for the SVM classifier contains more tweets labelled as others, the precision, recall and f-measure of categories: Prayer, Cash, Shelter, Rescue, and Relief were lower as compared to the category Others. Tweets labelled as cash may also be considered as relief or vice versa which often causes ambiguity in the data. Considering that the tweets retrieved were either Filipino or English, the system finds it hard to cluster similar disaster types in the Incremental Clustering Algorithm. This results to having data with similar disaster type be in several separate clusters.

For further improvements to the study, a different variation of Naïve Bayes algorithm could be used to improve the classification of Twitter data. A different approach can also be tested for better classification of disaster-related and unrelated tweets. A better similarity measure could also be used in the Incremental Clustering Algorithm to avoid having similar clusters be separated. A complete list of locations for the Name-Entity Recognition can improve the performance of Density-Based Spatiotemporal Clustering Algorithm in clustering data with similar locations. A different clustering algorithm can also be implemented to test if it is better than Density-Based Spatiotemporal Clustering Algorithm.

The training set used for SVM has a large number of tweets labelled as 'Others' giving the high f-measure score of 99% compared to the rest of the categories.

'Others' could be broken down into more specific need categories to avoid confusion on the assessment. An equal proportion of needs for the training data could improve the accuracy of SVM. For better classification of needs, 'relief' and 'cash' could also be combined since the most common words used in these categories are similar.

Algorithm. Retrieved April 7, 2017 from
http://ieeexplore.ieee.org/document/6988085